\documentclass[11pt,a4paper]{article}
\usepackage[hyperref]{naaclhlt2018}
\usepackage{times}
\usepackage{latexsym}
\usepackage{amssymb}
\usepackage{url}
\usepackage{enumitem}
\setlist{noitemsep}
\usepackage{multirow}
\usepackage{array}
\usepackage{graphicx}
\usepackage{fontenc}
\usepackage{textcomp}
\usepackage[english]{babel}

\aclfinalcopy 

\title{NLPR@SRPOL at SemEval-2019 Task 6 and Task 5: Linguistically enhanced deep learning offensive sentence classifier}

\author{Alessandro Seganti\textsuperscript{1}, Helena Sobol\textsuperscript{1}, Iryna Orlova\textsuperscript{1}, Hannam Kim\textsuperscript{2},\\ 
	{\bf Jakub Staniszewski\textsuperscript{1}, Tymoteusz Krumholc\textsuperscript{1},}
		{\bf Krystian Koziel\textsuperscript{1}} \\
\textsuperscript{1} Samsung R\&D Institute Poland \\
\textsuperscript{2} Samsung Electronics, Korea \\
	 {\small \tt a.seganti@samsung.com}}

\date{}

\begin{document}
\maketitle

\begin{abstract}
The paper presents a system developed for the SemEval-2019 competition Task 5 \textit{hatEval} \citet{Task5_description} (team name: \textit{LU Team}) and Task 6 \textit{OffensEval} \citet{offenseval} (team name: \textit{NLPR@SRPOL}), where we achieved $2^{nd}$ position in Subtask C. The system combines in an ensemble several models (LSTM, Transformer, OpenAI's GPT, Random forest, SVM) with various embeddings (custom, ELMo, fastText, Universal Encoder) together with additional linguistic features (number of blacklisted words, special characters, etc.). The system works with a multi-tier blacklist and a large corpus of crawled data, annotated for general offensiveness. In the paper we do an extensive analysis of our results and show how the combination of features and embedding affect the performance of the models.

\end{abstract}

\section{Introduction}

In 2017 two-thirds of all adults in the United States have experienced some form of online harassment \citep{PEW2017}.\footnote{Due to the topic of the SemEval-2019 Tasks 5 and 6, the present paper contains offensive expressions spelled out in full. These are solely illustrations of the problems under consideration. They should not be interpreted as expressing our views in any way.} This, together with various episodes of online harassment, boosted research on the general problem of recognizing and/or filtering offensive language on the Internet. Still, recognizing if a sentence expresses hate speech against immigrants or women, understanding if a sentence is offensive to a group of people, an individual or others -- these tasks continue to be very difficult for neural networks and machine learning models to accomplish. In order to do this, various implementations have been proposed; for the most successful recent approaches see \citet{TopSOTA,UnifiedDeepLearning,ExMachina,waseem-hovy:2016:N16-2,park2017one,HateSonar}. Most of them use various combination of features to recognize these characteristics.

This article presents a system that we have implemented for recognizing if a sentence is offensive. The system was developed for two SemEval-2019 competition tasks:
Task 5 \textit{hatEval} ``Multilingual detection of hate speech against immigrants and women in Twitter'' \citet{Task5_description} (team name: \textit{LU Team}) and Task 6 \textit{OffensEval} ``Identifying and categorizing offensive language in social media'' \citep{offenseval} (team name: \textit{NLPR@SRPOL}). Table \ref{semeval_results} shows the results that we achieved with our system in the SemEval-2019 competitions.

\begin{table}[h!]
	\centering
	\begin{tabular}{|c|c|} 
		\hline
		\textbf{Competition} & \textbf{Placement}	\\
		\hline
		Task 6-A &  $8^{th}$ position \\
		\hline
		Task 6-B & $9^{th}$ position  \\
		\hline      
		Task 6-C &  $2^{nd}$ position  \\
		\hline
		Task 5-A & $8^{th}$ position (ex aequo) \\ 
		\hline
	\end{tabular}
	\caption{SemEval-2019 results.}
	\label{semeval_results}
\end{table}

In order to create a highly accurate classifier, we combined state-of-the-art AI with linguistic findings on the pragmatic category of impoliteness \citep{Culpeper2011Politeness,JayJanchewitz,BrownLevinson}. We achieved this by deciding on the factors that point to the impoliteness of a given expression (for the blacklists) or the entire sentence (for corpus annotation). Such factors led us to divide the blacklist into ``offensive'' and ``offensive in context'', as most linguistic studies of impoliteness focus on various aspects of the context. Furthermore, linguistic research made it possible to arrive at a maximally general definition of offensiveness for the crowdsourced annotators.

The article is organized as follows. Section \ref{sec:SOTA} presents the current state of the art for offensive sentence classification. Section \ref{sec:system} explains the architecture of our system (features, models and ensembles). Section \ref{sec:data} describes the datasets and how they were created. Section \ref{sec:results} shows the results of the SemEval-2019 tasks in detail, motivating which combination of features and models was the best. Finally, section \ref{sec:conclusions} offers conclusions together with our plans for future research.

\section{Related work}
\label{sec:SOTA}
In recent years, the problem of recognizing if a sentence is offensive or not has become an important topic in the machine learning literature. The problem itself has different declinations depending on the point of view. Currently there are three main areas of research in this topic in the machine/deep learning community:
\begin{enumerate}
	\item Distinguishing offensive language from non-offensive language;
	\item Solving biases in deep learning systems;
	\item Recognizing more specific forms of offensivness (e.g. racism, sexism etc.).
\end{enumerate}
The main problem with each of the tasks is the amount of data available to researchers for experimenting with their systems. This -- together with the fact that it is difficult to clearly define what is offensive/racist/sexist or not -- makes the three problems listed above very difficult for a deep learning system to solve. 

Articles have showed that there is a strong bias in text and embeddings, and have tried to solve this bias using different techniques \citep{Zhao2018Learning, Dixon2018Measuring,ManIsToComputer}. Furthermore, thanks to a dataset defined in \citet{waseem-hovy:2016:N16-2} and \citet{waseem:2016:NLPandCSS}, various works have gone in the direction of recognizing sexism and racism in tweets \citep{TopSOTA, park2017one}.

Another field of work was recognizing offensiveness in the Wikipedia internal discussion forum dataset \citep{ExMachina}. This dataset has led to other articles making systems for distinguishing between offensive and non-offensive language \citep{UnifiedDeepLearning, TopSOTA, BenchmarkingAggression, AllYouNeedIsLove, RNNToxixComment, park2017one, ChallengesForToxic}.

Linguistic expertise enhanced the functionality at two stages: sentence annotation (described in detail in Section \ref{sec:data}) and active creation of blacklists (Section \ref{sec:system}). The completion of these tasks breaks new ground, as there exist no corpus linguistic studies on the generality of offensive language, to the best of our knowledge. Recent approaches of narrower scope are \citet{BritishBollocks} and \citet{SwearingInEnglish}.

\section{System description}
\label{sec:system}
Our system is composed of three major elements, described below:
\begin{itemize}
	\item Features -- common to all models;
	\item Various models -- neural networks or not;
	\item Ensemble.
\end{itemize}

\subsection{Features}
\label{sec:features}
This section describes the features that we used and explains their role.
We implemented the following features:
\begin{itemize}
	\item Number of blacklisted words in the sentence;
	\item Number of special characters, uppercase characters, etc.;
	\item A language model taught to recognize offensive and not offensive words.
\end{itemize}

\paragraph {Blacklisted} We used two kinds of blacklisted expressions: ``offensive'' and ``offensive in context''. The ``offensive in context'' expressions are offensive in specific contexts and unoffensive otherwise, e.g. \emph{bloody} or \emph{pearl necklace}. This dictionary was compiled by crowdsourcing and contains about 2,300 words (+ variations). The blacklist consists of swear words, invectives, profanities, slurs and other impolite expressions.

\paragraph {Special characters, uppercase, etc.} We checked the graphemic characteristics of the written text and we gave this as a feature to the model. We mainly used the non user related features defined in \citet{UnifiedDeepLearning}.

\paragraph {Language model} Inspired by the work of \citet{LMNER2018}, we decided to train a language model on both offensive and non-offensive words. For this purpose, we trained two character based language models, one on the offensive dictionary (described above) and the other from a corpus of non-offensive words. After training them we used the difference in perplexity of each input word as a feature for the model.

\subsection{Models}
We trained various models and then combined them in an ensemble. This section outlines the models that were part of the ensemble.

\paragraph {Embeddings} Both the Neural networks and the machine learning models used embeddings. We used the following embeddings: ELMo \citep{ELMo}, fastText \citep{fastText}, custom embeddings, and Universal Sentence Encoder \citep{UniversalSentenceEncoder}. For fastText, we used the 1 million word (300d) vectors trained on Wikipedia 2017, below called fastText 1M. 

The custom embedding was built by training a fastText embedding on our corpus. We then combined the 1M fastText embeddings with these custom embeddings using Truncated SVD after concatenating their columns (this was done inspired by the work \citep{ConceptNet2017}). Building custom embeddings was important for the offensive word classification because the original version of the fastText 1M embeddings contained around 50\% of the words in the corpus while after adding the custom embeddings, only 30\% of the words were out of the vocabulary. Below, this combination of embeddings is called ``combined''.

\paragraph {Neural networks} We used two types of neural network models:
\begin{itemize}
	\item LSTM models \citep{hochreiter1997lstm};
	\item Transformer models \citep{vaswani2017attention}.
\end{itemize}
For both models, we used multi-head attention and we tried different embeddings. In most cases, the Transformer models had better results than the LSTM models, and this is what we used in the submissions. The parameters of the models are described in Appendix \ref{sec:parameters}. In both models, the Features described in Section \ref{sec:features} are concatenated to the output of the model.

\paragraph {OpenAI GPT} One of the models that we used was the OpenAI GPT \citep{Finetune}. We used the GPT model in its original form, without changing any parameters. Our results show that this model works very well when there is enough data for finetuning. However, small classes -- as in Task 6 Subtask C -- pose a problem (see Section \ref{sec:results}).

\paragraph {Machine learning models} We used two machine learning models:
\begin{itemize}
	\item Random forest;
	\item SVM.
\end{itemize}
For these models, we built a pipeline where:
\begin{itemize}
	\item In a first step we either compute the embeddings of the sentences or get the Td-Idf score.
	\item In a second step we concatenate the result of the first step with the Features described in Section \ref{sec:features} (if used).
	\item We run the classifier.
\end{itemize}
As embeddings we used only the Universal Encoder, and with good results.

\paragraph {Ensemble} For the ensemble we used a voting classifier with soft voting (based on the probability returned by each model). For each subtask, we show which combination of models gave the best results.

\begin{figure}
\centering
\includegraphics[scale=0.35]{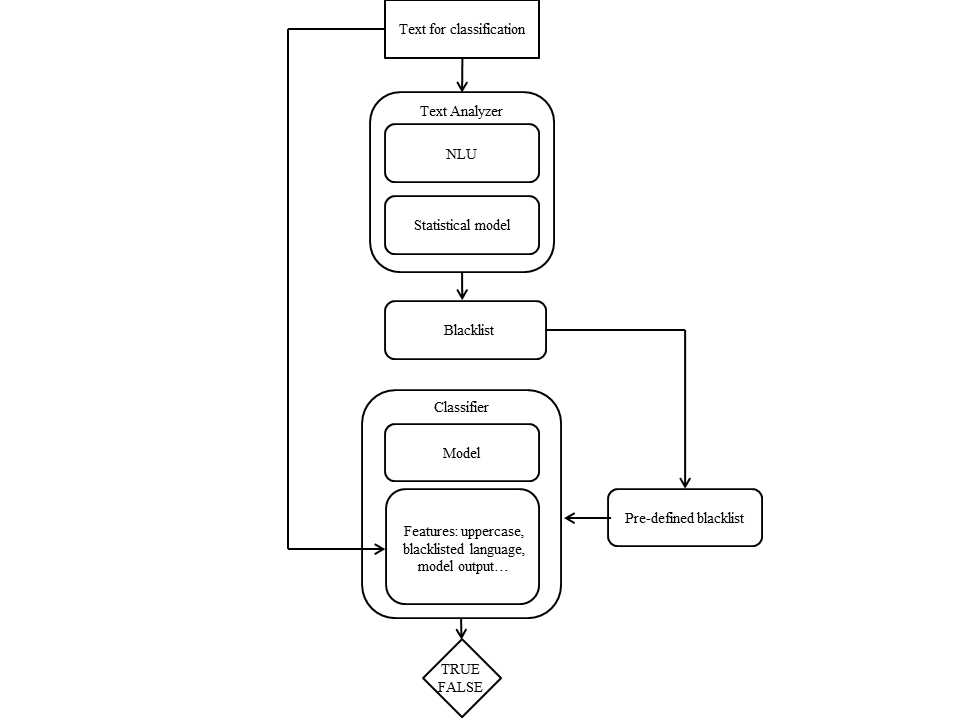}
\caption{Pipeline for the offensive sentence classifier.}
\label{fig:pipeline}
\end{figure}

The pipeline for the entire offensive sentence classifier is shown in Figure \ref{fig:pipeline}.

\section{Data/Datasets}
\label{sec:data}
\subsection{Preprocessing}
\label{sec:preprocessing}
Preprocessing plays a crucial role in the analysis of potentially offensive sentences, because most inputs use highly non-standard language. Hence, preprocessing was mainly focused on normalizing the language for simplifying the model work. We applied the following preprocessing:
\begin{itemize}
	\item Substituting user names with \textless USER\textgreater{} tokens;
	\item Removing all links;
	\item Normalizing words and letters;
	\item Normalizing spacing and non-standard characters;
	\item Over/Downsampling of the classes;
\end{itemize}

After the preprocessing, we split by space and used each token as an input to the models.

\paragraph {Normalizing words and letters} We have a dictionary containing common spelling variants of words found in our corpus. We used this to change words to the ``canonical'' form. Examples of such variants can be seen in Table \ref{spelling_variants}.

\begin{table}[h!]
\centering
\tabcolsep=0.11cm
\begin{tabular}{|c|c|} 
	\hline
	\textbf{Word} & \textbf{Common variants}	\\
	\hline
	fuck & fvck, fok, fucc, phuk \\
	\hline
	nigger & n1gga, n1gr, niigr, nuggah, nigg3r  \\
	\hline      
	boob &  booob, booooooob \\
	\hline
	motherfucker & Mutha Fukker, Motha Fuker \\ 
	\hline
	ass & a55, 455 (``leetspeak'' variants) \\
	\hline
	assclown & \H{a}\v{s}\d{s}\c{ć}\'{l}$\sigma$\.{w}$\eta$ (vulgarity obfuscation) \\
	\hline
\end{tabular}
\caption{Common spelling variants.}
\label{spelling_variants}
\end{table}

\paragraph{Over/Downsampling} For each Task/Subtask, we systematically oversampled the classes to obtain a balanced dataset. This was especially important for Task 6/Subtask C, which introduced 3 highly unbalanced classes. For most subtasks we did two things at the same time:
\begin{itemize}
	\item Downsampled the majority class when there was too much difference from the other classes;
	\item Oversampled the minority class after downsampling.
\end{itemize}

\subsection{Datasets}
\label{sec:datasets_building}
In this section we give a high level overview of the datasets we used for training our models for the SemEval-2019 tasks. Detailed statistics are presented in Appendix \ref{sec:appdata}.
For training the model, we used several openly available datasets:
\begin{itemize}
	\item \textit{Hate Sonar} gathered from Twitter \citep{HateSonar};
	\item 2 related hate speech datasets from Twitter \citep{waseem-hovy:2016:N16-2}, \citep{waseem:2016:NLPandCSS};
	\item Insulting internet comments \citep{Kaggle};
	\item Attacking, aggressive, toxic and neutral comments from Wikipedia Talk Pages \citep{ExMachina};
	\item \textit{Vulgar Twitter} \citep{VulgarTwitter};
\end{itemize}  
\noindent our own custom-built corpora and datasets provided by the SemEval organizers. 

From the sources listed above, we added a total of 20,399 sentences to the SemEval-2019 corpus for Task 5, and 97,759 sentences to the one for Task 6.

\paragraph{Custom Offensive language corpus} Our custom dataset was built by crowdsourcing and by crawling content from the Internet. The dataset is balanced, with 49,179 not offensive and 48,580 offensive comments. Around half of the dataset was labeled by linguists, who were asked to look for ``general offensiveness''. This could take various forms:
\begin{itemize}
	\item Expletives, swear-words, offensive terms;
	\item Rude meaning;
	\item Meaning that is harsh politically/ethically/ emotionally, and hence expression of hate/ disgust/disrespect;
	\item Uncomfortable topics related to the human genitals in a gross way;
	\item Hate speech, sarcasm, sexism, racism, violence, etc.;
	\item Discussion of drug use or other illegal actions;
	\item For any other reasons, children should not have access to the sentence.
\end{itemize}

To each sentence, the linguists assigned one of the three labels:
\begin{itemize}
	\item OFF -- offensive sentence,
	\item NOT -- not offensive sentence,
	\item Nonsense -- random collection of words or non-English (removed from the corpus).
\end{itemize}

In cases of disagreement between linguists, we chose the most popular label, if applicable, or obtained an expert annotation. We calculated Fleiss' kappa for inter-annotator agreement \citep{Fleiss1971}, which extends Cohen's kappa to more than two raters \citep{Cohen1960}. For random ratings Fleiss' $\kappa = 0$, while for perfect agreement $\kappa = 1$. Our $\kappa$ was equal to 0.62, which falls in the ``substantial agreement'' category, according to \citet{LandisKoch}.

The remaining part of the corpus was assessed automatically with a blacklist-based filter. 

\paragraph{Dataset for Task 6} The OLID dataset \citep{OLID} contains Offensive and Not Offensive sentences. The Offensive sentences are further categorized into:
\begin{itemize}
	\item TIN -- targeted insults and threats,
	\item UNT -- untargeted.
\end{itemize}
and the targeted (TIN) category was further subdivided into:
\begin{itemize}
	\item IND -- individual target,
	\item GRP -- group target,
	\item OTH -- a target that is neither an individual nor a group.
\end{itemize}

Our full offensive language corpus, described in the previous subsection, was used for this task. The OFF sentences were further annotated for the two categories while the NOT sentences were not further annotated. All the additional classes were added automatically by a wordlist-based annotator.

\paragraph{Dataset for Task 5} 
The dataset for Task 5 \citep{Task5_description} contained the classes:
\begin{itemize}
	\item HATE -- hate speech against women or immigrants,
	\item NOHATE -- no hate speech against women or immigrants.
\end{itemize}
 together with other subclasses. Given that we participated in the Task 5 Subtask A, we annotated our corpus only with these two labels. Using a mixture of automated and manual annotation, we were able to add around 30k sentences from our dataset for this task.

\section{Results}
\label{sec:results}
\paragraph{SemEval}
In Table \ref{ensemble_detailed_results_semeval} we show the average F1 of our models for all the SemEval-2019 Tasks and Subtasks. These results were obtained by using an ensemble of models and in Table \ref{what_ensemble_contain_table} we show which model was used inside which ensemble. The acronyms used in the table correspond to:
\begin{itemize}
	\item \textbf{GPT} : OpenAI's GPT model
	\item \textbf{RF}: Random Forest
	\item \textbf{T}: Transformer model
	\item \textbf{U}: Universal encoder
	\item \textbf{EL}: ELMo embeddings
	\item \textbf{CO}: Combined embeddings (see \ref{sec:features} for an explanation of this)
	\item \textbf{F}: Features.
\end{itemize} 

\begin{table}[h!]
	\centering
	\begin{tabular}{|c|c|} 
		\hline
		\textbf{Competition} & \textbf{Macro F1}	\\
		\hline
		Task 6-A & 0.80 \\
		\hline
		Task 6-B & 0.69  \\
		\hline      
		Task 6-C & 0.63 \\
		\hline
		Task 5-A & 0.51 \\ 
		\hline
	\end{tabular}
	\caption{SemEval-2019 results breakdown.}
	\label{ensemble_detailed_results_semeval}
\end{table}

\begin{table}[h!]
	\centering
	\begin{tabular}{|c|c|c|c|c|}
		\hline
		\textbf{Models} & \textbf{6-A} & \textbf{6-B} & \textbf{6-C} & \textbf{5-A}	\\
		\hline
		GPT & \checkmark &  & \checkmark* & \checkmark \\
		\hline
		RF  & \checkmark & \checkmark & \checkmark & \checkmark \\
		\hline
		RF+U & \checkmark & \checkmark && \checkmark \\
		\hline
		T+EL  & \checkmark &   & \checkmark & \checkmark  \\
		\hline
		T+CO+U & \checkmark &  & \checkmark* & \checkmark \\
		\hline
		T+EL+U+F  &  &  &  & \checkmark \\
		\hline
	\end{tabular}
	\caption{Ensemble detail. The models marked with * have been trained with an unbalanced dataset.}
	\label{what_ensemble_contain_table}
\end{table}

Given the short amount of time, during the SemEval competition we were unable to test all the combinations of models and data preparation types to choose the best combination for the Ensemble. We thus selected the models in the ensemble by experimenting with part of the models. This is the main reason why only one model used in Task 5 contains additional features (the TELUF model). 

After the competition, we tried the models contained in the Ensembles on all the tasks; detailed results are presented in Table \ref{table:result_all_models_all_tasks} in the Appendix. It is important to note though that the results in the Appendix cannot be directly compared with the ones of the SemEval competition because although the models were the same, the Test data was different (the golden data has not been released yet).

From the results we clearly see that we have two ``data regimes'': in the \textit{low data regime} (Task 6 Subtask B and C), Random forest (with or without the Universal embeddings) is the best choice. However, in \textit{the big(ger) data regime}, Fine tune is the best model. Also in the low data regime each model works best with a different data preparation strategy: GPT with unbalanced data, the Transformer with oversampled and downsampled data while Random forest with oversampled data.

\paragraph{Ablation studies}
In this part we show the results of ablation studies on the transformer and random forest models. In this study, we want to understand how far the final result was influenced by the linguistically based features and preprocessing we defined in this article. All the results obtained in this section have been computed on a Test set created from the Train set shared in the SemEval-2019 tasks (as in Appendix \ref{sec-appendix:model_results}). As we discussed in the previous section, the Tasks were characterized by a ``low'' (Task 6 C) and a ``big'' data regime (Task 6 A), thus we compare the ablation study results for these two extreme regimes.

\begin{table}
	\centering
	\begin{tabular}{|c|c|c|}
		\hline
		\textbf{Model} & \textbf{Task 6 A} & \textbf{Task 6 C} \\
		\hline
		T + CO & 0.73 & 0.44 \\
		\hline
		T + CO + U & 0.71 & 0.52 \\
		\hline
		T + CO + F & \textbf{0.75} & 0.45 \\
		\hline
		T + CO + U + F & 0.74 & 0.47 \\ 
		\hline
		RF & 0.7 & \textbf{0.54} \\
		\hline
		RF + F & 0.68 & 0.43 \\
		\hline
		RF + U & 0.72 & 0.48 \\
		\hline
		RF + U + F & 0.69 & 0.38 \\
		\hline
	\end{tabular}
	\caption{Macro F1 for selected Transformer models with different combinations of features}
	\label{table:ablation_studies}
\end{table}

In a first study we wanted to understand how the features influenced the results. For this reason, we tried some combinations of Features, Embeddings and Models on both Task 6 Subtask A and Task 6 Subtask C; the relevant macro F1 results are shown in Table \ref{table:ablation_studies}. The table shows that, in the big data regime, the Random Forest works best when only the Universal Encoder is used, while the Transformer model improves its performance when the features are added. On the other hand, in the low data regime, we see that the plain Random Forest outperforms all the other combinations. This is probably because the more things we add, the more the model needs to learn, and with little data this is simply not possible.

In a second study we wanted to understand how much the normalization defined in Section \ref{sec:preprocessing} affected the performance of the model. For this reason, we trained again the best models in Table \ref{table:ablation_studies} for both Subtasks with an unnormalized version of the dataset. The results are that for Subtask A, the model \textbf{T + CO + F} F1 decreased from 0.75 to 0.73 while for Subtask C, the \textbf{RF} F1 decreased from 0.54 to 0.44. 

The results of this section seem to point to the fact that the features we added and the normalization we used are beneficial for the performance of the models. Further work will be devoted to understanding this point though.

\section{Conclusions}
\label{sec:conclusions}
The article presented our approach to making a classifier recognizing offensive expressions in text. It showed how our architecture is suitable for multiple (related) offensive sentence classification tasks. It also showed how we built the features and the data that the model used for learning. Thanks to our system, we were $2^{nd}$ in the SemEval-2019 Task 6/Subtask C. In the article we also showed with ablation studies that the linguistic features proposed and the embeddings added improve the performance of the models we used.

In the future, we will extend our system to recognize a wider set of features. We are currently working on analyzing the linguistic differences between the offensive corpus and the non-offensive corpus. Specifically, we think that by analyzing the differences, we should be able to build a ``white-list'' of terms that can be used as features that will help the classifier understand which sentences are less likely to be offensive.

\section*{Acknowledgments}
We would like to thank G. Knor, P. Przybysz, P. Andruszkiewicz, P. Bujnowski and most of the AI Team in Samsung R\&D Institute Poland for all the helpful discussions that made this article possible.

\bibliography{NLPR_SRPOL_Semeval_Submission}

\begin{thebibliography}{36}
\expandafter\ifx\csname natexlab\endcsname\relax\def\natexlab#1{#1}\fi

\bibitem[{Aken et~al.(2018)Aken, Risch, Krestel, and
  L{\"o}ser}]{ChallengesForToxic}
Betty~van Aken, Julian Risch, Ralf Krestel, and Alexander L{\"o}ser. 2018.
\newblock \href {https://arxiv.org/abs/1809.07572} {Challenges for toxic
  comment classification: An in-depth error analysis}.
\newblock arXiv:1809.07572.

\bibitem[{Basile et~al.(2019)Basile, Bosco, Fersini, Nozza, Patti, Rangel,
  Rosso, and Sanguinetti}]{Task5_description}
Valerio Basile, Cristina Bosco, Elisabetta Fersini, Debora Nozza, Viviana
  Patti, Francisco Rangel, Paolo Rosso, and Manuela Sanguinetti. 2019.
\newblock Semeval-2019 task 5: Multilingual detection of hate speech against
  immigrants and women in twitter.
\newblock In \emph{Proceedings of the 13th International Workshop on Semantic
  Evaluation (SemEval-2019)}. Association for Computational Linguistics”,
  location = “Minneapolis, Minnesota.

\bibitem[{Bojanowski et~al.(2017)Bojanowski, Grave, Joulin, and
  Mikolov}]{fastText}
Piotr Bojanowski, Edouard Grave, Armand Joulin, and Tomas Mikolov. 2017.
\newblock \href {http://aclweb.org/anthology/Q17-1010} {Enriching word vectors
  with subword information}.
\newblock \emph{Transactions of the Association for Computational Linguistics},
  5:135--146.

\bibitem[{Bolukbasi et~al.(2016)Bolukbasi, Chang, Zou, Saligrama, and
  Kalai}]{ManIsToComputer}
Tolga Bolukbasi, Kai-Wei Chang, James~Y. Zou, Venkatesh Saligrama, and Adam~T.
  Kalai. 2016.
\newblock \href {https://arxiv.org/pdf/1607.06520.pdf} {Man is to computer
  programmer as woman is to homemaker? {D}ebiasing word embeddings}.
\newblock In \emph{Advances in Neural Information Processing Systems}, pages
  4349--4357.

\bibitem[{Brown and Levinson(1987)}]{BrownLevinson}
Penelope Brown and Stephen~C. Levinson. 1987.
\newblock \emph{Politeness: Some Universals in Language Usage}.

\bibitem[{Cachola et~al.(2018)Cachola, Holgate, Preoţiuc-Pietro, and
  Li}]{VulgarTwitter}
Isabel Cachola, Eric Holgate, Daniel Preoţiuc-Pietro, and Junyi~Jessy Li.
  2018.
\newblock \href {http://aclweb.org/anthology/C18-1248} {Expressively vulgar:
  The socio-dynamics of vulgarity and its effects on sentiment analysis in
  social media}.
\newblock In \emph{Proceedings of the $27^{th}$ International Conference on
  Computational Linguistics}, pages 2927--2938.

\bibitem[{Cer et~al.(2018)Cer, Yang, Kong, Hua, Limtiaco, St.~John, Constant,
  Guajardo-Cespedes, Yuan, Tar, Sung, Strope, and
  Kurzweil}]{UniversalSentenceEncoder}
Daniel Cer, Yinfei Yang, Sheng-yi Kong, Nan Hua, Nicole Limtiaco, Rhomni
  St.~John, Noah Constant, Mario Guajardo-Cespedes, Steve Yuan, Chris Tar,
  Yun-Hsuan Sung, Brian Strope, and Ray Kurzweil. 2018.
\newblock \href {https://arxiv.org/abs/1803.11175} {Universal {S}entence
  {E}ncoder}.
\newblock arXiv:1803.11175.

\bibitem[{Cohen(1960)}]{Cohen1960}
Jacob Cohen. 1960.
\newblock \href {https://journals.sagepub.com/doi/10.1177/001316446002000104}
  {A coefficient of agreement for nominal scales}.
\newblock \emph{Educational and Psychological Measurement}, 20(1):37--46.

\bibitem[{Culpeper(2011)}]{Culpeper2011Politeness}
Jonathan Culpeper. 2011.
\newblock Politeness and impoliteness.
\newblock In \emph{Pragmatics of Society}, pages 391--436.

\bibitem[{Davidson et~al.(2017)Davidson, Warmsley, Macy, and Weber}]{HateSonar}
Thomas Davidson, Dana Warmsley, Michael Macy, and Ingmar Weber. 2017.
\newblock \href
  {https://aaai.org/ocs/index.php/ICWSM/ICWSM17/paper/view/15665/14843}
  {Automated hate speech detection and the problem of offensive language}.
\newblock In \emph{Proceedings of the Eleventh International AAAI Conference on
  Web and Social Media (ICWSM 2017)}, pages 512--515.

\bibitem[{Dewaele(2015)}]{BritishBollocks}
Jean-Marc Dewaele. 2015.
\newblock \href {https://core.ac.uk/download/pdf/42134101.pdf} {British
  \emph{bollocks} versus {A}merican \emph{jerk}: Do native {B}ritish {E}nglish
  speakers swear more -- or differently -- compared to {A}merican {E}nglish
  speakers?}
\newblock \emph{Applied Linguistics Review}, 6(3):309--339.

\bibitem[{Dixon et~al.()Dixon, Li, Sorensen, Thain, and
  Vasserman}]{Dixon2018Measuring}
Lucas Dixon, John Li, Jeffrey Sorensen, Nithum Thain, and Lucy Vasserman.
\newblock \href
  {http://www.aies-conference.com/wp-content/papers/main/AIES_2018_paper_9.pdf}
  {Measuring and mitigating unintended bias in text classification}.
\newblock In \emph{Proceedings of the AAAI/ACM Conference on AI, Ethics, and
  Society}.

\bibitem[{Duggan(2017)}]{PEW2017}
Maeve Duggan. 2017.
\newblock \href
  {http://www.pewinternet.org/wp-content/uploads/sites/9/2017/07/PI_2017.07.11_Online-Harassment_FINAL.pdf}
  {Online harassment 2017}.

\bibitem[{Fleiss(1971)}]{Fleiss1971}
Joseph~L. Fleiss. 1971.
\newblock \href
  {http://www.wpic.pitt.edu/research/biometrics/Publications/Biometrics%20Archives%20PDF/395-1971%20Fleiss0001.pdf}
  {Measuring nominal scale agreement among many raters}.
\newblock \emph{Psychological Bulletin}, 76(5):378--382.

\bibitem[{Founta et~al.(2018)Founta, Chatzakou, Kourtellis, Blackburn, Vakali,
  and Leontiadis}]{UnifiedDeepLearning}
Antigoni-Maria Founta, Despoina Chatzakou, Nicolas Kourtellis, Jeremy
  Blackburn, Athina Vakali, and Ilias Leontiadis. 2018.
\newblock \href {https://arxiv.org/abs/1802.00385} {A unified deep learning
  architecture for abuse detection}.
\newblock arXiv:1802.00385.

\bibitem[{Gr{\"o}ndahl et~al.(2018)Gr{\"o}ndahl, Pajola, Juuti, Conti, and
  Asokan}]{AllYouNeedIsLove}
Tommi Gr{\"o}ndahl, Luca Pajola, Mika Juuti, Mauro Conti, and N.~Asokan. 2018.
\newblock \href {https://arxiv.org/pdf/1808.09115.pdf} {All you need is
  ``love'': Evading hate speech detection}.
\newblock arXiv:1808.09115.

\bibitem[{Hochreiter and Schmidhuber(1997)}]{hochreiter1997lstm}
Sepp Hochreiter and J{\"u}rgen Schmidhuber. 1997.
\newblock \href
  {http://citeseerx.ist.psu.edu/viewdoc/download?doi=10.1.1.676.4320&rep=rep1&type=pdf}
  {Long {S}hort-{T}erm {M}emory}.
\newblock \emph{Neural computation}, 9(8):1735--1780.

\bibitem[{Impermium(2012)}]{Kaggle}
Impermium. 2012.
\newblock \href
  {https://www.kaggle.com/c/detecting-insults-in-social-commentary/data}
  {Detecting insults in social commentary}.

\bibitem[{Jay and Janschewitz(2008)}]{JayJanchewitz}
Timothy Jay and Kristin Janschewitz. 2008.
\newblock \href
  {https://www.mcla.edu/Assets/MCLA-Files/Academics/Undergraduate/Psychology/Pragmaticsofswearing.pdf}
  {The pragmatics of swearing}.
\newblock \emph{Journal of Politeness Research}, 4:267--288.

\bibitem[{Kumar et~al.(2018)Kumar, Ojha, Malmasi, and
  Zampieri}]{BenchmarkingAggression}
Ritesh Kumar, Atul~Kr. Ojha, Shervin Malmasi, and Marcos Zampieri. 2018.
\newblock \href {http://aclweb.org/anthology/W18-4401} {Benchmarking aggression
  identification in social media}.
\newblock In \emph{Proceedings of the First Workshop on Trolling, Aggression
  and Cyberbullying}, pages 1--11.

\bibitem[{Landis and Koch(1977)}]{LandisKoch}
J.~Richard Landis and Gary~G. Koch. 1977.
\newblock \href
  {https://www.jstor.org/stable/2529310?seq=1#page_scan_tab_contents} {The
  measurement of observer agreement for categorical data}.
\newblock \emph{Biometrics}, 33(1):159--174.

\bibitem[{Li(2018)}]{RNNToxixComment}
Siyuan Li. 2018.
\newblock \href {https://escholarship.org/uc/item/5f87h061} {Application of
  recurrent neural networks in toxic comment classification}.
\newblock Master's thesis, University of California, Los Angeles.

\bibitem[{McEnery(2006)}]{SwearingInEnglish}
Tony McEnery. 2006.
\newblock \emph{Swearing in English. Bad language, purity and power from 1586
  to the present}.
\newblock Routledge, London and New York.

\bibitem[{Park and Fung(2017)}]{park2017one}
Ji~Ho Park and Pascale Fung. 2017.
\newblock \href {https://arxiv.org/abs/1706.01206} {One-step and two-step
  classification for abusive language detection on {T}witter}.
\newblock arXiv:1706.01206.

\bibitem[{Peters et~al.(2018)Peters, Neumann, Iyyer, Gardner, Clark, Lee, and
  Zettlemoyer}]{ELMo}
Matthew~E. Peters, Mark Neumann, Mohit Iyyer, Matt Gardner, Christopher Clark,
  Kenton Lee, and Luke Zettlemoyer. 2018.
\newblock \href {http://www.aclweb.org/anthology/N18-1202} {Deep contextualized
  word representations}.
\newblock In \emph{Proceedings of NAACL-HLT 2018}, pages 2227--2237.

\bibitem[{Pitsilis et~al.(2018)Pitsilis, Ramampiaro, and Langseth}]{TopSOTA}
Georgios~K. Pitsilis, Heri Ramampiaro, and Helge Langseth. 2018.
\newblock \href {https://arxiv.org/pdf/1801.04433} {Detecting offensive
  language in tweets using deep learning}.
\newblock arXiv:1801.04433.

\bibitem[{Radford et~al.(2018)Radford, Narasimhan, Salimans, and
  Sutskever}]{Finetune}
Alec Radford, Karthik Narasimhan, Tim Salimans, and Ilya Sutskever. 2018.
\newblock \href
  {https://www.cs.ubc.ca/~amuham01/LING530/papers/radford2018improving.pdf}
  {Improving language understanding by generative pre-training}.
\newblock Technical report, OpenAI.

\bibitem[{Speer et~al.(2017)Speer, Chin, and Havasi}]{ConceptNet2017}
Robert Speer, Joshua Chin, and Catherine Havasi. 2017.
\newblock \href
  {http://www.aaai.org/ocs/index.php/AAAI/AAAI17/paper/download/14972/14051}
  {Conceptnet 5.5: An open multilingual graph of general knowledge}.
\newblock In \emph{Proceedings of the Thirty-First AAAI Conference on
  Artificial Intelligence (AAAI-17)}, pages 4444--4451.

\bibitem[{Vaswani et~al.(2017)Vaswani, Shazeer, Parmar, Uszkoreit, Jones,
  Gomez, Kaiser, and Polosukhin}]{vaswani2017attention}
Ashish Vaswani, Noam Shazeer, Niki Parmar, Jakob Uszkoreit, Llion Jones,
  Aidan~N Gomez, {\L}ukasz Kaiser, and Illia Polosukhin. 2017.
\newblock \href
  {https://papers.nips.cc/paper/7181-attention-is-all-you-need.pdf} {Attention
  is all you need}.
\newblock In \emph{Advances in Neural Information Processing Systems}, pages
  5998--6008.

\bibitem[{Waseem(2016)}]{waseem:2016:NLPandCSS}
Zeerak Waseem. 2016.
\newblock \href {http://aclweb.org/anthology/W16-5618} {Are you a racist or am
  {I} seeing things? {A}nnotator influence on hate speech detection on
  {T}witter}.
\newblock In \emph{Proceedings of the First Workshop on NLP and Computational
  Social Science}, pages 138--142.

\bibitem[{Waseem and Hovy(2016)}]{waseem-hovy:2016:N16-2}
Zeerak Waseem and Dirk Hovy. 2016.
\newblock \href {http://www.aclweb.org/anthology/N16-2013} {Hateful symbols or
  hateful people? {P}redictive features for hate speech detection on
  {T}witter}.
\newblock In \emph{Proceedings of the NAACL Student Research Workshop}, pages
  88--93.

\bibitem[{Wulczyn et~al.(2017)Wulczyn, Thain, and Dixon}]{ExMachina}
Ellery Wulczyn, Nithum Thain, and Lucas Dixon. 2017.
\newblock \href {https://arxiv.org/abs/1610.08914} {Ex machina: Personal
  attacks seen at scale}.
\newblock In \emph{Proceedings of the $26^{th}$ International Conference on
  World Wide Web}, pages 1391--1399.

\bibitem[{Yu et~al.(2018)Yu, Mayhew, Sammons, and Roth}]{LMNER2018}
Xiaodong Yu, Stephen Mayhew, Mark Sammons, and Dan Roth. 2018.
\newblock \href {https://arxiv.org/pdf/1809.05157.pdf} {On the strength of
  character language models for multilingual named entity recognition}.
\newblock In \emph{Proceedings of the 2018 conference on empirical natural
  language processing (EMNLP 2018)}.

\bibitem[{Zampieri et~al.(2019{\natexlab{a}})Zampieri, Malmasi, Nakov,
  Rosenthal, Farra, and Kumar}]{OLID}
Marcos Zampieri, Shervin Malmasi, Preslav Nakov, Sara Rosenthal, Noura Farra,
  and Ritesh Kumar. 2019{\natexlab{a}}.
\newblock {Predicting the Type and Target of Offensive Posts in Social Media}.
\newblock In \emph{Proceedings of NAACL}.

\bibitem[{Zampieri et~al.(2019{\natexlab{b}})Zampieri, Malmasi, Nakov,
  Rosenthal, Farra, and Kumar}]{offenseval}
Marcos Zampieri, Shervin Malmasi, Preslav Nakov, Sara Rosenthal, Noura Farra,
  and Ritesh Kumar. 2019{\natexlab{b}}.
\newblock {SemEval-2019 Task 6: Identifying and Categorizing Offensive Language
  in Social Media (OffensEval)}.
\newblock In \emph{Proceedings of The 13th International Workshop on Semantic
  Evaluation (SemEval)}.

\bibitem[{Zhao et~al.(2018)Zhao, Zhou, Li, Wang, and Chang}]{Zhao2018Learning}
Jieyu Zhao, Yichao Zhou, Zeyu Li, Wei Wang, and Kai-Wei Chang. 2018.
\newblock \href {https://arxiv.org/pdf/1809.01496.pdf} {Learning gender-neutral
  word embeddings}.
\newblock In \emph{Proceedings of the 2018 Conference on Empirical Methods in
  Natural Language Processing}, pages 4847--4853.

\end{thebibliography}
\bibliographystyle{acl_natbib}

\clearpage

\appendix

\section{Supplementary Materials}
\subsection{Model parameters}
\label{sec:parameters}
The following parameters were used for all LSTM and Transformer models in the results Section \ref{sec:results}:
\begin{itemize}
	\item keep probability: 0.8;
	\item LSTM units: 100;
	\item L2 regularization: 0;
	\item fully connected size: 256 or 128;
	\item multihead attention:
		\begin{itemize}
			\item attention size: 5 or 2, 
			\item attention head: 4
		\end{itemize}
\end{itemize}
For GPT, we used a learning rate of 6.25e-5 and an L2 regularization of 0.01.

\subsection{Data}
\label{sec:appdata}

In Table \ref{table:overall_data_statistics} we show the amount of data that was contained in our corpus (overall). In Table \ref{table:task_5_data_statistics} and \ref{table:task_6_data_statistics} we show the data for Task 5 and Task 6. For a description of how these corpora were built and annotated, see Section \ref{sec:datasets_building}. 

\begin{table}[h!]
\centering
\begin{tabular}{|c|r|r|r|} 
	\hline
	\textbf{Source} & \textbf{NOT} & \textbf{OFF} & \textbf{Total} \\ 
	\hline
	Custom corpus & 16,545 & 12,938 & 29,483\\
	\hline
	Kaggle & 2,629 & 3,463 & 6,092\\
	\hline
	Twitter & 917 & 23,438 & 24,355\\
	\hline
	Wikipedia & 29,088 & 8,741 & 37,829\\
	\hline
	\textbf{Total} & 49,179 & 48,580 & 97,759\\
	\hline
\end{tabular}
\caption{Statistics for our offensive language corpus. The Kaggle dataset was collected by \citet{Kaggle}. The Twitter dataset was compiled from 4 sources: \citet{HateSonar}, \citet{VulgarTwitter}, \citet{waseem-hovy:2016:N16-2} and \citet{waseem:2016:NLPandCSS}. The Wikipedia dataset was collected by \citet{ExMachina}.}
\label{table:overall_data_statistics}
\end{table}

\begin{table}[h!]
\centering
\begin{tabular}{|c|r|} 
	\hline
	\textbf{Class} & \textbf{Total} \\
	\hline
	HATE & 16,508  \\
	\hline
	NOHATE & 11,154 \\
	\hline
\end{tabular}
\caption{Statistics for the additional corpus for SemEval-2019 Task 5.}
\label{table:task_5_data_statistics}
\end{table}

\begin{table}[h!]
\centering

\begin{tabular}{|c|c|c|r|} 
	\hline
	\textbf{Class} & \textbf{Targeting} & \textbf{Target} & \textbf{Total}  \\
	\hline
	\multirow{3}{*}{OFF} & \multirow{4}{*}{TIN} & IND & 18,506  \\
	\cline{3-4}
	& & GRP & 6,761 \\
	\cline{3-4}
	& & OTH & 1,025 \\
	\cline{3-4}
	&   & Total & 34,669 \\
	\cline{2-4} 
	& UNT  & -- & 6,234 \\
	\cline{2-4}      
	& Total & -- &  59,837  \\
	\hline 
	NOT & -- & -- & 64,773 \\
    \hline
\end{tabular}
\caption{Statistics for the additional corpus for SemEval-2019 Task 6.}
\label{table:task_6_data_statistics}
\end{table}

\subsection{Model results}
\label{sec-appendix:model_results}
In this section we show the detailed results of all the models for all the SemEval-2019 tasks. For each Task, we extracted a test set from the Train data released by SemEval. We compared the models to one of the current state of the art defined in \citet{park2017one}; the results shown here are obtained by averaging the best F1 for each class (not a single model). The data by \citet{waseem-hovy:2016:N16-2} for comparing to the state-of-the-art model has been kindly shared by the authors of \citet{park2017one}. In the table we marked with 
\begin{itemize}
	\item No additional mark: the normalized data with oversampling and downsampling as described in Section \ref{sec:data}.
	\item \textbf{FULL}: the normalized data with oversampling but no downsampling.
	\item \textbf{UNB}: the normalized data without oversampling or downsampling.
\end{itemize}
The model acronyms are the same as the ones used in Section \ref{sec:results}.

\newpage
\begin{table*}[h!]
	\centering
	\begin{tabular}{|c|c|c|c|c|c|c|c|c|}
		\hline
		\textbf{Model} & \textbf{5-A} & \textbf{6-A} & \textbf{6-B} & \textbf{6-B FULL} & \textbf{6-C} & \textbf{6-C FULL} & \textbf{6-C UNB} & \textbf{SOTA} \\
		\hline
		RF & 0.7 & 0.62 & \textbf{0.61} & 0.58 & 0.44 & \textbf{0.54} & 0.45 & 0.78 \\
		\hline
		RF + F & 0.68 & 0.68 & 0.59 & 0.54 & 0.32 & 0.43 & 0.41 & - \\
		\hline
		RF + U & 0.72 & 0.69 & 0.6 & 0.55 & 0.39 & 0.48 & 0.46 & 0.74 \\
		\hline
		GPT & \textbf{0.77} & \textbf{0.77} & 0.58 & \textbf{0.6} & 0.42 & 0.49 & \textbf{0.51} & \textbf{0.81} \\
		\hline
		T + CO + U & 0.74 & 0.71 & 0.58 & \textbf{0.6} & \textbf{0.52} & 0.45 & 0.5 &  0.73 \\
		\hline
		T + EL & 0.73 & 0.73 & 0.58 & 0.58 & 0.49 & 0.5 & 0.45 & 0.74 \\
		\hline
		SOTA & - & - & - & - & - & - & - & 0.78 \\
		\hline
	\end{tabular}	
	\caption{Macro F1 for all the models on all the Tasks and on the state-of-the-art (SOTA) data.}	
	\label{table:result_all_models_all_tasks}
\end{table*}

\end{document}